\documentclass[letterpaper, 10 pt, conference]{ieeeconf}  

\IEEEoverridecommandlockouts                              

\usepackage[linesnumbered,ruled,vlined]{algorithm2e}

\usepackage{graphicx,adjustbox}
\usepackage[caption=false,font=footnotesize,lofdepth,lotdepth]{subfig}

\usepackage{cite}
\usepackage{tabularx}
\usepackage{makecell}

\usepackage{amsmath}
\usepackage{amssymb}
\usepackage{amsfonts}
\usepackage{color}
\usepackage{url}
\usepackage{soul}
\usepackage[bookmarks=true]{hyperref}
\usepackage[usenames,x11names]{xcolor}
\usepackage{xspace} 
\usepackage{multicol}
\usepackage[normalem]{ulem}

\usepackage{multirow} 
\usepackage{tikz,pgf}
\usetikzlibrary{arrows,automata,shapes,calc,backgrounds,spy,positioning,shadows}

\usepackage{soul}
\setul{1pt}{.4pt}

\graphicspath{{figures/}}

\SetKw{Break}{break}

\newtheorem{theorem}{Theorem}[section]

\newtheorem{definition}{Definition}[section]
\newtheorem{lemma}[theorem]{Lemma}

\newtheorem{example}{Example}[section]

\newtheorem{problem}{Problem}[section]

\renewcommand\baselinestretch{0.98}\selectfont

\newcommand{\CA}[1]{\mathcal{#1}}
\newcommand{\BF}[1]{\mathbf{#1}}
\newcommand{\BB}[1]{\mathbb{#1}}

\newcommand{\BS}[1]{\boldsymbol{#1}}

\newcommand{\Next}{\BF{X}}
\newcommand{\Event}{\BF{F}}
\newcommand{\Until}{\CA{U}}
\newcommand{\True}{\top}

\newcommand{\AP}{{AP}}
\newcommand{\Pref}{{R}}

\newcommand{\PA}{\mathcal{A}}

\newcommand{\FA}{{\mathcal{A}_\phi}}
\newcommand{\TS}{\mathcal{T}}

\newcommand{\ES}{\mathcal{E}}

\newcommand{\card}[1]{\left| {#1} \right|}
\newcommand{\spow}[1]{2^{#1}}

\newcommand{\dto}{\rightrightarrows}

\graphicspath{ {./final_images/} }

\renewcommand{\baselinestretch}{0.98}

\title{\LARGE \bf
A$^*$-based Temporal Logic Path Planning with User Preferences on Relaxed Task Satisfaction
}

\author{Disha Kamale, Xi Yu, Cristian-Ioan Vasile
\thanks{*This work was not supported by any organization}
\thanks{Disha Kamale and Cristian-Ioan Vasile are with the Mechanical Engineering and Mechanics Department at Lehigh University, PA, USA, {\tt\small \{ddk320, cvasile\}@lehigh.edu}}%
\thanks{Xi Yu is with the School of Manufacturing Systems and Networks at Arizona State University, AZ, USA, {\tt\small xyu@asu.edu}}%
}

\begin{document}

\maketitle
\thispagestyle{empty}
\pagestyle{empty}

\begin{abstract}
In this work, we consider the problem of planning for temporal logic tasks in large robot environments. When full task compliance is unattainable, we aim to achieve the best possible task satisfaction by integrating user preferences for relaxation into the planning process. 
Utilizing the automata-based representations for temporal logic goals and user preferences, we propose an A$^*$-based planning framework. This approach effectively tackles large-scale problems while generating near-optimal high-level trajectories. To facilitate this, we propose a simple, efficient heuristic that allows for planning over large robot environments in a fraction of time and search memory as compared to uninformed search algorithms. We present extensive case studies to demonstrate the scalability, runtime analysis as well as empirical bounds on the suboptimality of the proposed heuristic.  
\end{abstract}

\section{{Introduction}}
\label{sec:intro}

With the rapidly growing integration of robots into real-world applications, the need for time-efficient, sophisticated frameworks for successfully executing complex tasks is increasingly prominent. In this work, we consider the problem of planning for tasks expressed as temporal logic (TL) goals. TL formulations are particularly valuable due to their rich semantics, which enable precise articulation of complex requirements for robotic systems~\cite{fainekos2005hybrid, vasile2013sampling, lindemann2019robust, smith2011optimal}. Traditionally, the problem of planning for TL specifications is approached using automata-based, sampling-based, optimization-based or learning-based techniques~\cite{vasile2017minimum, kamale2021automata,cardona2022partial, leung2023backpropagation, cai2023overcoming}.  

In this work, we are interested in designing a fast, scalable path-planning framework over large environments for given temporal logic specifications. A critical challenge in TL planning is that failing to meet even a minor sub-requirement can render the entire task infeasible. In such scenarios, it becomes crucial to still achieve meaningful satisfaction of the task as closely as possible.
To facilitate this, we incorporate user preferences for relaxation of specifications into the planning framework. In the literature, various notions of relaxation including \textit{maximizing probability of satisfaction~\cite{lahijanian2011temporal, rahmani2023probabilistic}}, \textit{deadline relaxation}\cite{vasile2017time}, \textit{minimum revision}, \textit{minimum violation}~\cite{vasile2017minimum}, \textit{partial satisfaction}~\cite{cardona2022partial, amorese2023optimal}. These methods often employ automata-based techniques, constructing explicit product automata for graph search to find optimal high-level trajectories. While this approach provides a clear notion of progress towards satisfaction, its scalability is limited when dealing with large environments or complex tasks. 
 
For large robot environments, complex task specifications, and large number of user preferences, the runtime of path planning can rapidly increase. Several works consider the path planning problems over large environments utilizing techniques ranging from contraction hierarchies\cite{wang2021constrained}, sampling-based methods~\cite{vasile2013sampling, kantaros2020stylus} to hierarchical planning. Informed search algorithms such as A$^*$ have been found useful at efficiently solving these large-scale problems over discrete search spaces~\cite{likhachev2003ara, amorese2023optimal, khalidi, bhattacharya2010multi}. However, the efficiency of A$^*$ depends largely on the heuristic function, which provides an estimate of the cost to reach the goal. The worst-case time and memory complexity of A$^*$ is exponential in depth of search.
To address this, several variants, such as the weighted A$^*$~\cite{pohl1970heuristic, felner2003kbfs}, have been proposed to enhance search performance, though at the expense of optimality guarantees~\cite{ebendt2009weighted}. 

We propose a heuristic function for TL task planning that efficiently reduces the number of nodes explored to find a solution by leveraging the problem's structure.
The primary objective of this work is to develop a fast
planning approach for robots deployed in large environments with syntactically co-safe Linear Temporal Logic (scLTL) tasks and user preferences on relaxed satisfaction in case of infeasibility.
Similar to our previous works~\cite{kamale2021automata, kamale2024optimal}, we represent the user preferences as a weighted-finite-state-edit system.
By leveraging the abstractions to encapsulate robot motion, specification and user preferences, we propose a heuristic-based
path planning framework.
We trade-off optimality guarantees in planning for
search efficiency, measured by the reduction in explored nodes and improved runtime performance.

This work differs from closely related works~\cite{kamale2021automata, kamale2024optimal, amorese2023optimal, khalidi} in several aspects. In~\cite{kamale2021automata}, we considered an explicit product automaton construction to handle multiple notions of relaxations. As opposed to the optimization-based approach in~\cite{kamale2024optimal}, this work considers a heuristic-based search method. In~\cite{khalidi}, the authors address the problem of TL planning, without allowing relaxations to the specification.  In~\cite{amorese2023optimal}, the authors present an efficient A$^*$-based approach to address TL planning with partial satisfaction. On the contrary, we consider multiple notions of relaxations such as minimum revision problem (MRP), minimum violation problem (MVP). 

The main contributions of this work are threefold: 
1) We propose a heuristic-based planning algorithm for temporal logic tasks and user preferences for relaxation that achieves near-optimal trajectories.
2) We propose an efficient heuristic based on progress in the relaxed specification automaton which captures the specification and all user-preferred relaxations in case of infeasible sub-specifications
3) Our extensive case studies demonstrate the efficacy of the proposed heuristic in terms of a significant improvement in memory and computation time across various examples. Moreover, we present the runtime analysis of the proposed heuristic with respect to different components of TL planning problem.  


\noindent \textbf{Notation} The symbols $\mathbb{R}$, $\mathbb{Z}$, and $\mathbb{B}$ represent the sets of real, integer, and binary numbers respectively. The set of integers greater than or equal to $a$ is denoted by $\mathbb{Z}_{\geq a}$. For a set $X$, $2^X$ and $|X|$ denote its power set and cardinality, respectively. If $\Sigma$ is an alphabet, then $\Sigma^*$ represents the language consisting of all finite words over $\Sigma$.

\section{Problem Setup}

In this section, we formally introduce the problem of temporal logic path planning with relaxation. We begin with a detailed description of models expressing the robot's motion in the environment, the temporal logic task description as well as formal model encapsulating the user's preferences for relaxation in case the original specification is infeasible. 

\subsection{Robot and Environment Model}

We consider a robot deployed in a fully known planar environment within which the robot can move deterministically. The environment may contain multiple labeled regions. We consider a finite abstraction of the robot's motion in the environment as a weighted transition system, a widely followed approach in formal control synthesis~\cite{vasile2017minimum, baier2008principles, wongpiromsarn2012receding}.

\begin{definition}[Transition System]
A weighted transition system (TS) is a tuple
$\TS = (X, x^\TS_0, \delta_\TS, \AP, \ell, w_\TS)$, where 
$X$ is a finite set of states indicating regions in the environment; 
$x^\TS_0 \in X$ is the initial state;
$\delta_\TS \subseteq X \times X$ is a set of transitions which captures the set of permissible robot movements in the environment;
$\AP$ is a set of labels (atomic propositions);
$\ell \colon X \to {\AP}$ is a labeling function;
$w_\TS \colon \delta_\TS \to \BB{R}_{\geq 0}$ is a weight function.
\end{definition}

Note that in addition to transitions between regions, $\delta_\TS$ also contains self-loop transitions, allowing the robot to stay stationary at any state $x\in X$. The weight function $w_\TS(x_{k},x_{k+1})$ represents path length from $x_k$ to $x_{k+1}$. Naturally, traversing a self-loop incurs a cost of $0$. 

As the robot moves through the environment, it generates a (potentially infinite) sequence of states $\BF{x} = x_0, x_{1} \ldots$, referred to as a \emph{trajectory} (or run) of a robot, such that $(x_{k}, x_{ k+1}) \in \delta_\TS$ for all $k \in \mathbb{Z}_{\geq 0}$ and $x_0 = x^\TS_0$.
When the robot is at a state $x$ labeled with $\pi \in \AP$, the atomic proposition $\pi$ is said to be $\mathsf{true}$. 
The set of all trajectories of $\TS$ is $Runs(\TS)$.
A state trajectory $\BF{x}$ generates an \emph{output trajectory} $\BF{o} = o_0 o_1 \ldots$,
where $o_k = h(x_k)$ for all $k \geq 0$.
We also denote an output trajectory by $\BF{o}=\ell(\BF{x})$.
The {\em (generated) language} corresponding to a TS $\TS$ is
the set of all generated output words, which we denote by $\CA{L}(\TS)$.
We define the weight of a trajectory as
$w_\TS(\BF{x}) = \sum_{k=1}^{\card{\BF{x}}} w_\TS(x_{k-1}, x_k)$.

\subsection{Temporal Logic Specification}
In this work, we use syntactically co-safe Linear Temporal Logic (scLTL) to formally define the robot tasks.
scLTL composes symbols from $\Sigma=\spow{\AP}$ with logical and temporal operators with the following syntax:
\begin{align}
    \varphi :=  \True \; \vert \;\pi \; \vert \; \neg \varphi \; \vert \; \varphi \land  \varphi \; \vert\;\varphi \;\Until \; \varphi \;|\; \Next \varphi 
    \label{eq:syntax_scltl}
\end{align}
where,  $\pi \in \AP$, $\True$ denotes logical $\mathsf{true}$ value, \textit{{negation}} ($\neg$) and \textit{conjunction} ($\land$) are Boolean operators while \textit{next} ($\Next$) and \textit{until} ($\Until$) denote the temporal operators. Additional Boolean operators such as \textit{disjunction} ($\lor$) and temporal operators such as  \textit{eventually} ($\Event$) can be derived using~\ref{eq:syntax_scltl}. 
Intuitively, the formula $\Next \varphi$ denotes that $\varphi$ holds true at the next step, $\Event \varphi$ indicates that $\varphi$ is satisfied at some point in the future, and $\varphi_1 \Until \varphi_2$
denotes that $\varphi_1$ is true until $\varphi_2$ becomes true. For a detailed description of the syntax and semantics of scLTL, we refer the reader to~\cite{baier2008principles}. 

Although the semantics of scLTL formulae is defined over infinite words, such as the ones produced by $\TS$, its satisfaction can be decided in finite time enabled by {\textit{finite good prefixes}~\cite{kupferman2001model}}. An scLTL formula is said to be satisfied by trajectory $\mathbf{x}$, denoted as $\mathbf{x} \models \phi$, if and only if the resulting output word satisfies  $\phi$, i.e., $\mathbf{o} \models \phi$.

\subsection{User preferences for relaxation}

If part of the task specification $\phi$ should become infeasible, the user relaxation preferences facilitate meaningful satisfaction of the task. 

\begin{definition}[User Relaxation Preference]
\label{def:task-preference}
Let $L$ be a language over the alphabet $2^\AP$.
A \emph{user task preference} is a pair $(\Pref, w_\Pref)$,
where $\Pref\subseteq L\times(2^\AP)^*$ is a relation that
captures how words in $L$ can be transformed to words from $(2^\AP)^*$ and is of the form $\sigma \mapsto^p \sigma'$ where $\sigma \in L, \sigma' \in (\spow{AP})^*$. $w_\Pref \colon  \Pref \to \BB{R}$ represents
the cost of the word transformations. The relation $\Pref$ can also be understood as a multi-valued
function $\Pref\colon L \dto (2^\AP)^*$.
\end{definition}

\begin{example}
\label{ex:spec}
Consider the task of picking up bread and ice-cream from a supermarket. The scLTL specification is $\phi = (\Event \; {bread_{s}})  \land  \Event \; {ice\_cream_{s}}$, where ${bread_{s}}$ and ${ice\_cream_{s}}$ are atomic propositions. If infeasible, the user-preferred relaxation options are 1) picking up bread from the nearest local bakery for a penalty of 5, 2) picking up ice cream from the nearest shop for a penalty of 7, 3) removing ice-cream from the list for a penalty of 12. The formal representation for these preferences is $1) bread_{s} \mapsto^5 bread_{bakery}, 2)ice\_cream_{s} \mapsto^7 ice\_cream_{shop}, 3) ice\_cream_{s} \mapsto^{12} \epsilon $. Note that (1) and (2) are instances of MRP while (3) is an instance of MVP. 
\end{example}

\subsection{Problem definition}
The problem of optimal temporal logic path planning with relaxation is defined as follows.

\begin{problem} [Optimal TL Path Planning with \allowdisplaybreaks Relaxations] 
Given the robot motion abstraction as a weighted transition system $\TS$, an scLTL task specification $\phi$ as well as user preferences for relaxation ($\Pref, w_\Pref$), 
find a path $\mathbf{\Tilde{x}}$ in $\TS$ that satisfies the specification $\phi$ while only performing necessary relaxations
and minimizes the cost of the trajectory.
Formally, 
\begin{gather*}
\label{prob:relaxed_planning}
    \min_{\BF{x}} \hat{J}(\BF{x}) = {w_\TS}(\BF{x}) + \lambda \cdot w_\Pref (\BF{o}, \BF{o}^{relax})\\
      \text{s.t. } \;  \BF{x}_0 = x^\TS_0 \quad \hfill{\text{(Initial state)}}, \\
     \BF{o} \models \phi \quad \hfill{\text{(Task satisfaction)}}, \\
     \BF{o}^{relax} = \ell(\BF{x}), \;
      (\BF{o}, \BF{{o}}^{relax}) \in R  \quad \hfill{\text{(Relaxation preferences)}} 
\end{gather*}
\end{problem}
where $\lambda$ represents a blending parameter, $\BF{o}^{relax} = \ell(\BF{{x}})$ denotes the output word of the $\TS$, and $\BF{o} \in (2^\AP)^*$ is a word satisfying $\phi$.

\section{Approach}


 In this section, we elaborate on our approach to Problem~\ref{prob:relaxed_planning} which involves several key steps. Given the task specification and user preferences for relaxation, we begin by converting them into automata. This facilitates the construction of a \textit{Relaxed Satisfaction Automaton}, introduced in~\cite{kamale2024optimal}, which efficiently encapsulates the original task specification and all user-preferred relaxations along with their associated penalties. The primary advantage this offers is that the progress toward task satisfaction can be explicitly considered even in the case of infeasible sub-specifications. Building upon this representation, we present the A$^*$-based search algorithm for temporal logic planning with relaxations, which is the main contribution of this work. Finally we present a discussion on practical considerations regarding the design of the heuristic for TL planning. 

\subsection{Automata models for temporal logic } 

  Any scLTL formula $\phi$ can be translated~\cite{duret.16.atva2} into a deterministic \textit{finite state automaton} (FSA) which accepts the set of all good prefixes of all words that satisfy $\phi$~\cite{kupferman2001model}, defined as follows.  

\begin{definition}[Deterministic Finite State Automaton]
\label{def:dfa}
A deterministic finite state automaton (DFA) is a tuple
$\FA = (S_\FA, s_0^\FA, \Sigma, \delta_\FA, F_\FA)$, where:
$S_\FA$ is a finite set of states;
$s^\FA_0 \in S_\FA$ is the initial state;
$\Sigma$ is the input alphabet;
$\delta_\FA \colon S_\FA \times \Sigma \to S_\FA$ is the transition function;
$F_\FA \subseteq S_\FA$ is the set of accepting states.

\end{definition}

A trajectory of the DFA $\BF{s} = s_0 s_1 \ldots s_{n+1}$ is generated by
a finite sequence of symbols $\BS{\sigma} = \sigma_0 \sigma_1 \ldots \sigma_n$
where $s_0 = s^\FA_0$ is the initial state of $\FA$ and
$s_{k+1} = \delta_\FA(s_k, \sigma_k)$
for all $0 \leq k \leq n$. This trajectory is said to be accepting if $s_{k+1} \in F_\FA$. 
The {\em (accepted) language} of a DFA $\FA$ is
the set of accepted input words denoted by $\CA{L}(\FA)$. Thus, in order to ensure the satisfaction of a formula $\phi$ by a trajectory $\BF{x}$ in environment $\TS$, it is necessary that $\BF{o}=\ell(\BF{x}) \in \mathcal{L}_\FA$.

In order to handle infeasibilities in $\phi$, the user preferences for relaxation are converted into a weighted finite state edit system defined as follows: 
\begin{definition}[Weighted Finite State Edit System]
\label{def:wfses}
A weighted finite state edit system (WFSE)
is a weighted DFA $\ES = (Z_\ES, z^\ES_0, \Sigma_\ES, \delta_\ES, F_\ES, w_\ES)$, where 
$\Sigma_\ES = \left(2^\AP\cup \{\epsilon\}\right) \times \left(2^\AP\cup \{\epsilon\}\right) \setminus \{(\epsilon, \epsilon)\}$,
$\epsilon$ denotes a missing or deleted symbol,
and $w_\ES\colon \delta_\ES \to \BB{R}$ is the transition weight function.
\end{definition}

The alphabet $\Sigma_\ES$ captures word edit operations such as addition, substitution, or deletion of symbols.
A transition $z' = \delta_\ES(z,(\sigma, \sigma'))$
{\color{black}
has input, output symbols $\sigma$ and $\sigma'$.
}

\noindent \textbf{Remark: }We add pass-through transitions to the WFSE with 0 weights to ensure that the satisfaction is not relaxed if the entire original specification is feasible. 

\begin{example}[Continuation of Example~\ref{ex:spec}] 
Consider states $z,z' \in Z_\ES$. Preference 1 can be captured by states $z,z'$ such that $z' \in (z, \delta_\ES(\sigma, \sigma'))$ where $\sigma$ is the input symbol such that $bread_{s} \in \sigma$, $\sigma'$ denotes the output symbol $bread_{bakery}\in \sigma'$ and $w_\ES(z, (\sigma, \sigma'), z') = 5$.  
\end{example}

\subsection{Relaxed Specification Product Automaton}

All permissible relaxations of the scLTL specification $\phi$ are captured using a product non-deterministic finite state automaton~\cite{kamale2024optimal} between the DFA $\FA$ and the WFSE $\ES$.

\begin{definition}[Relaxed Specification Automaton]
    Given a specification DFA $\FA = (S_\FA, s_0^\FA, \Sigma, \delta_\FA, F_\FA)$, the user task preferences represented as a WFSE $\ES = (Z_\ES, z^\ES_0, \Sigma_\ES, \delta_\ES, F_\ES, w_\ES)$, the relaxed specification automaton is 
    a tuple $\PA = (Q_{\PA}, q_{\PA}^0, \Sigma_\PA, \delta_{\PA}, F_{\mathcal{A}}, w_\PA)$, where $Q_\PA = Z_\ES \times S_\FA$ represents the state space; $q^\PA_0 = (z^\ES_0, s_0^\FA)$ is the initial state; 
        $\Sigma_\PA = \Sigma_\FA$ denotes the alphabet; 
        $\delta_\PA \subseteq Q_\PA \times \Sigma_\PA \times Q_\PA$ is a transition relation; 
        $F_\PA = F_\ES \times F_\FA \subseteq Q_\PA$ represents the set of final (accepting) states;
        $w_\PA : \delta_\PA \rightarrow \mathbb{R}_{\geq 0}$  is the weight function where $w_\PA(q, q') = w_\ES(z,z')$.
\end{definition}

We direct the reader to \cite{kamale2024optimal} for the relaxed satisfaction automaton construction.

\subsection{Heuristic-based for TL planning with relaxation}
For A$^*$ search on an arbitrary graph, the cost of node $n$ is given as $f(n) = g(n) + h(n)$. Thus, the cost computation utilizes two components of information: 1) the distance already covered to reach node $n$ from some initial node $n_0$, and 2) an estimated cost to reach a final node $n_f$ from node $n$. The function $f(\cdot)$ is often referred to as the \textit{f-score} of node $n$. The search continues by exploring the nodes with best \textit{f-scores} until $n_f$ is reached. The solution obtained by A$^*$ is guaranteed to be optimal if the heuristic is an underestimation of the actual cost to reach the goal for all nodes. The informativeness of the heuristic directly impacts the efficiency of the search.

For TL planning, designing such an informative heuristic is challenging. Unlike the traditional graph search problems where the final node $n_f$ is given, temporal logic tasks may require visiting multiple regions in the environment, where each visit to a location containing a desired label (atomic proposition) results in progress towards task satisfaction in the specification automaton. Potentially, the same atomic proposition could be satisfied at multiple locations in the environment. Thus, the state in the environment that results into reaching a final state in the specification automaton cannot be uniquely specified.

\noindent\textbf{Heuristic} We consider the distance to satisfaction in the relaxed satisfaction automaton scaled by a factor $\gamma$. Since $\PA$ contains multiple final states $q \in F_\PA$, we add a virtual final state $q^{\bowtie}$ to $\PA$ with only incoming edges $\{(q, q^{\bowtie}) \vert q \in F_\PA\}$. The heuristic value for node $(x,q)$ is, 
\begin{align}
    h(x,q) = \gamma \cdot d_{min} (q, q^{\bowtie} )
    \label{eq:heuristic}
\end{align}
where $d_{min}(q,q')$ denotes the minimum distance between nodes $q$ and $q'$ in $\PA$. The search is performed over the solution space consisting of $\TS$ and $\PA$. Note that, we avoid the explicit product construction and instead only enumerate the reachable states in this space by keeping track of labels of $\TS$ and the resulting transitions in $\PA$. Given a sequence of states $((x_0, q_0), (x_1,q_1), \ldots, (x_k, q_k))$, the graph distance for node $(x_{k+1}, q_{k+1})$ is
{\small
\begin{equation}
\label{eq:g_score}
\begin{aligned}
    g(x_{k+1},q_{k+1}) = \Sigma_{0}^{i=k-1} w_\TS(x_i, x_{i+1}) + 
    \lambda \cdot w_\PA(q_i, q_{i+1}) 
\end{aligned}
\end{equation}}%

Alg.~\ref{alg:A* search} outlines the search algorithm for TL planning with relaxation. Given the input $\TS, \FA$ and $\ES$, we construct the relaxed product automaton using the algorithm from~\cite{kamale2024optimal} (line 2). The initial node $(x_0^\TS, q_\PA^0)$ is added to the $queue$ to be processed. For node $(x,q)$, the \textit{f-score} is given by $f(x,q) = g(x,q) + h(x,q)$ where $g(x,q)$ is the graph distance of node $x$ from the initial node in $\TS$ and a graph distance in $\FA$ as in~\eqref{eq:g_score}
and $h(x,q)$ is obtained from~\eqref{eq:heuristic}. The $queue$ keeps track of the \textit{f-score}, current node, graph distances, and parent nodes. Given the current state $(x,q)$, all neighbors of $x$ in $\TS$ and the corresponding generated input symbols inform the next states $q'$ in $\PA$ leading to a new state $(x',q')$ in the solution space (lines 10-11). If unexplored, this node is investigated for graph score $\bar{g}$ and heuristic value. If the current path to $(x,q)$ from source is shorter than a previously stored path to $(x,q)$, we update the $g_{score}$ set. The $queue$ is updated with node $(x,q)$ and the associated information. 
Upon reaching the final state in $\PA$, the search stops and the resulting trajectory in the solution space is obtained using the stored parent nodes. Finally the $path$ is projected onto the robot environment to obtain the trajectory in the robot environment, denoted by $\textbf{x} = path_{\perp_\TS}$ (lines 6-8), where $\perp(\cdot)$ is the projection operator with $\perp(x_k, q_k) = x_k$.  
If the final state in $\PA$ cannot be reached, the routine returns infeasibility. 

\begin{algorithm}
\caption{\textit{A$^*$-based search for TL relaxation()}}
\label{alg:A* search}
\scriptsize{
\KwIn{$\TS, \phi, \ES, \gamma$}
\KwResult{$\BF{x}$}
\DontPrintSemicolon
\BlankLine
Initialize $queue \gets \emptyset$, $explored \gets \emptyset, g\_scores \gets \emptyset$\; 
$\PA$ = construct\_$\PA$($\ES, \FA$) \hfill{using~\cite{kamale2024optimal}} \;
$ g\_scores \gets \{(x_0^\TS, q_\PA^0) = 0\}, queue \gets (0, (x_0^\TS, q_\PA^0), 0, None)$\; 
\While{$queue \neq \emptyset$}{
$(x,q) \gets queue.pop()$\; 
\If{$q \in F_\PA$}{
reconstruct $path = ((x_0^\TS, q_\PA^0), \ldots, (x,q))$
return $path_{\perp_\TS}$
}
\lIf{$(x, q)$ in $explored$ }{
continue
}
$explored \gets (x, q)$ \;
\ForAll{$(x,x') \in \delta_\TS$}{
    \ForAll{$(q, (\ell(x'),\sigma'), q') \in \delta_\PA$}{
    \If{$(x',q')$ in $explored$ }{continue}
    $\bar g \gets g\_{scores}(x,q) + w_\TS(x,x') + \lambda \cdot w_\PA(q, q')$\hfill{ \ref{eq:g_score}} \; 
    $h(x',q') \gets get\_heuristic(q,q')$ \hfill{ \ref{eq:heuristic}}\; 
    \If{$(x',q') \in g\_scores$}{
    \lIf{$g\_scores(x',q') < \bar g$}{continue}
    \lElse{$g\_scores(x',q') \gets \bar g $}  
    }
    $queue \gets ((\bar g + h(q_\PA,q'), (x',q'), \bar g, (x,q)))$
    }
    }
 return $Infeasible$}
}
\end{algorithm}

\begin{lemma}[correctness]
The solution given by the proposed search algorithm always satisfies the specification. i.e., $ \BF{\Tilde{o}}^{relax} = h(\tilde{\BF{x}}) \models \phi$. \\
\begin{proof}Follows by construction and the stopping criterion the Alg.~\ref{alg:A* search}. 
\end{proof}
\end{lemma}




\subsection{Heuristic discussion}
The explicit product computation scales multi-linearly with $|\TS|$, $|\FA|$ and $|\ES|$ restricting the scalability due to the exhaustive search of the reachable space for solutions. By implicitly enumerating the states in the solution space, the proposed search algorithm Alg.~\ref{alg:A* search} offers a significant improvement in terms of runtime and memory even with zero heuristic. 

\begin{figure}[h]
\vspace{-4 pt}
    \centering
    \includegraphics[width=0.8\linewidth]{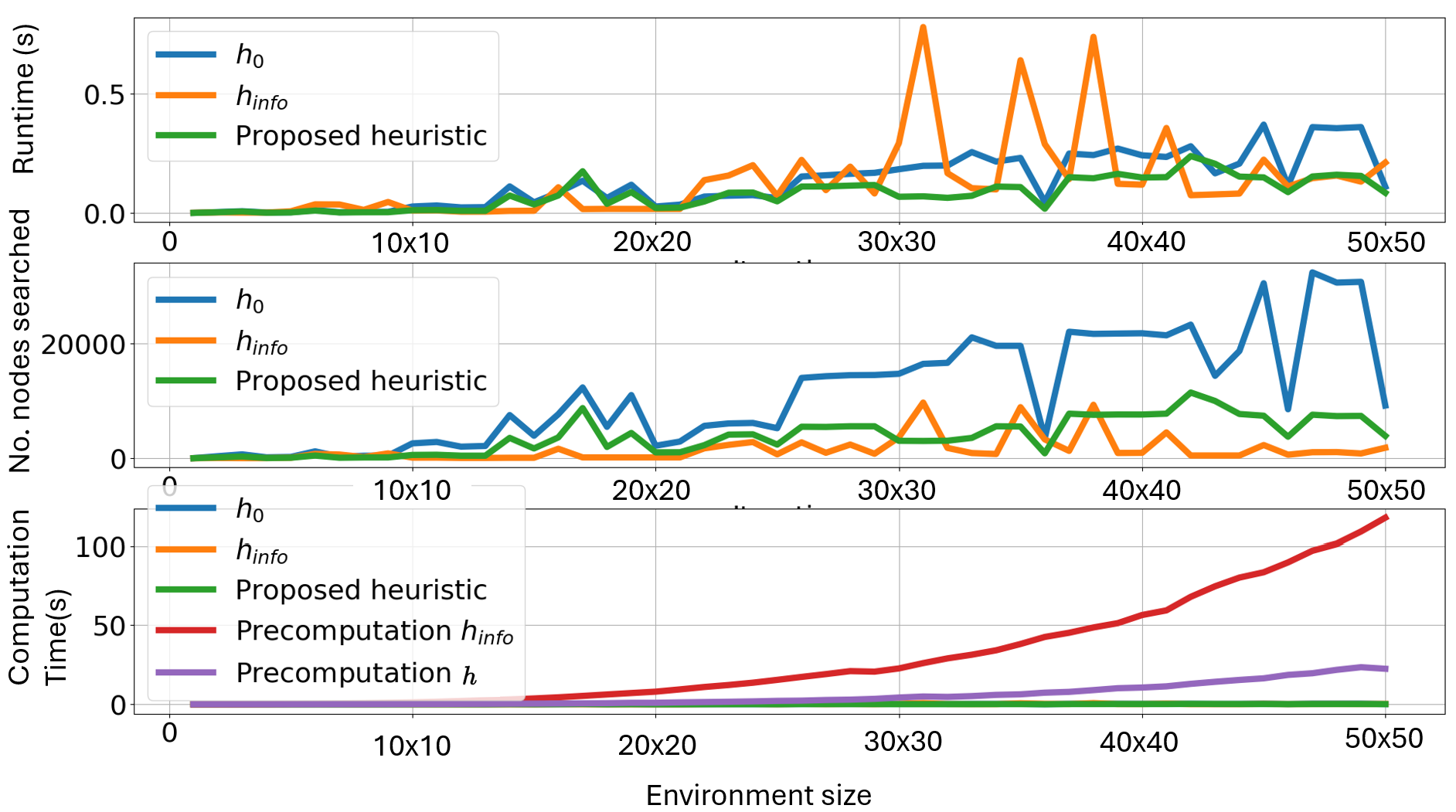}
    \caption{\scriptsize{Runtime, memory, and precomputation time for $h=0, h_{info}$ and proposed heuristic $h$. }}
    \vspace{-2pt}
    \label{fig:all-heuristics-comparison}
\end{figure}

Further improvement in search performance can be achieved by carefully crafting a heuristic function 
that balances informativeness and computational efficiency.
While a highly informative heuristic can reduce the number of explored nodes, it may worsen overall performance due to increased computational cost compared to simple node expansion.
For instance, consider the runtime and search results in Fig.~\ref{fig:all-heuristics-comparison}. $h_{info}$ refers to an informative heuristic which takes into account the symbol $\sigma$ needed to transition closer to the final state in $\PA$ and computes the distance to one of the nodes in the environment that contains this symbol. As shown in Fig.~\ref{fig:all-heuristics-comparison}, even though $h_{info}$ reduces the number of nodes explored, it significantly increases the runtime due to complex computations involved at each step. A commonly followed approach to circumvent this problem involves precomputing the necessary values. With slight changes to the problem setting, repeating these precomputations can quickly become impractical for large scale environments. 





\begin{figure*}[t]
    \centering
    \includegraphics[width=.75\linewidth]{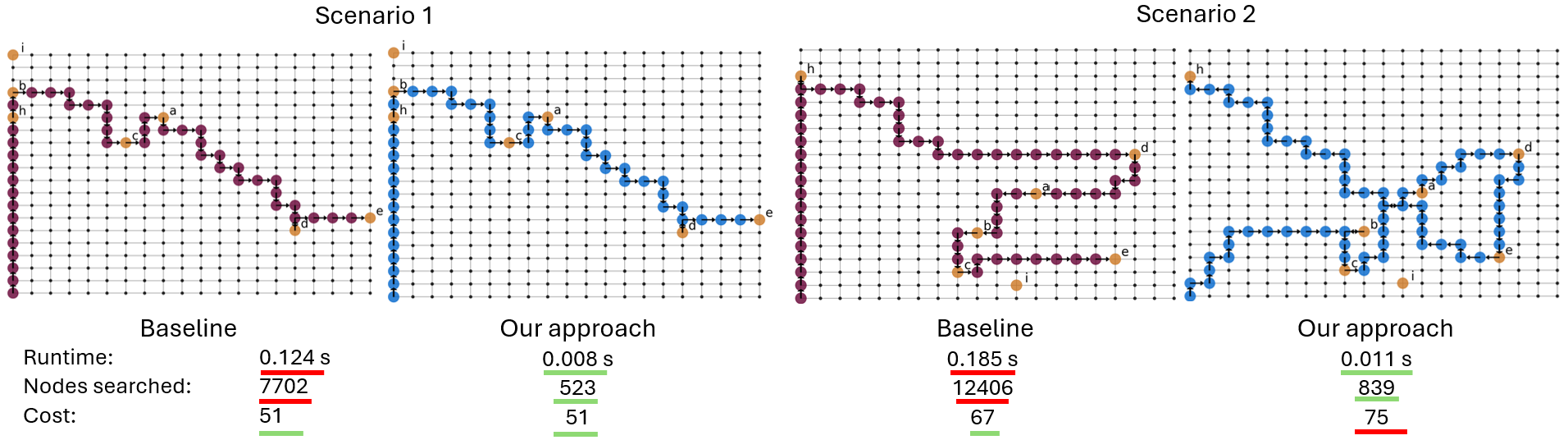}
    \caption{\scriptsize{Trajectories obtained using baseline and the proposed heuristic. Nodes with atomic propositions of interest are shown in yellow.}}
    \label{fig:cs1}
    \vspace{-4pt}
\end{figure*}
\section{Case studies}

In this section, we show the functionality of the proposed search algorithm, a comparison with a baseline uninformed search, and a runtime analysis. These studies were performed on Dell Precision 3640 Intel i9 with 64 GB RAM using Python 3.9.7. For all following studies, we set $\lambda=1$. 

\subsection{ Functionality} 

\begin{figure}[h]
    \centering
    \vspace{-5pt}
    \includegraphics[width=0.75\linewidth]{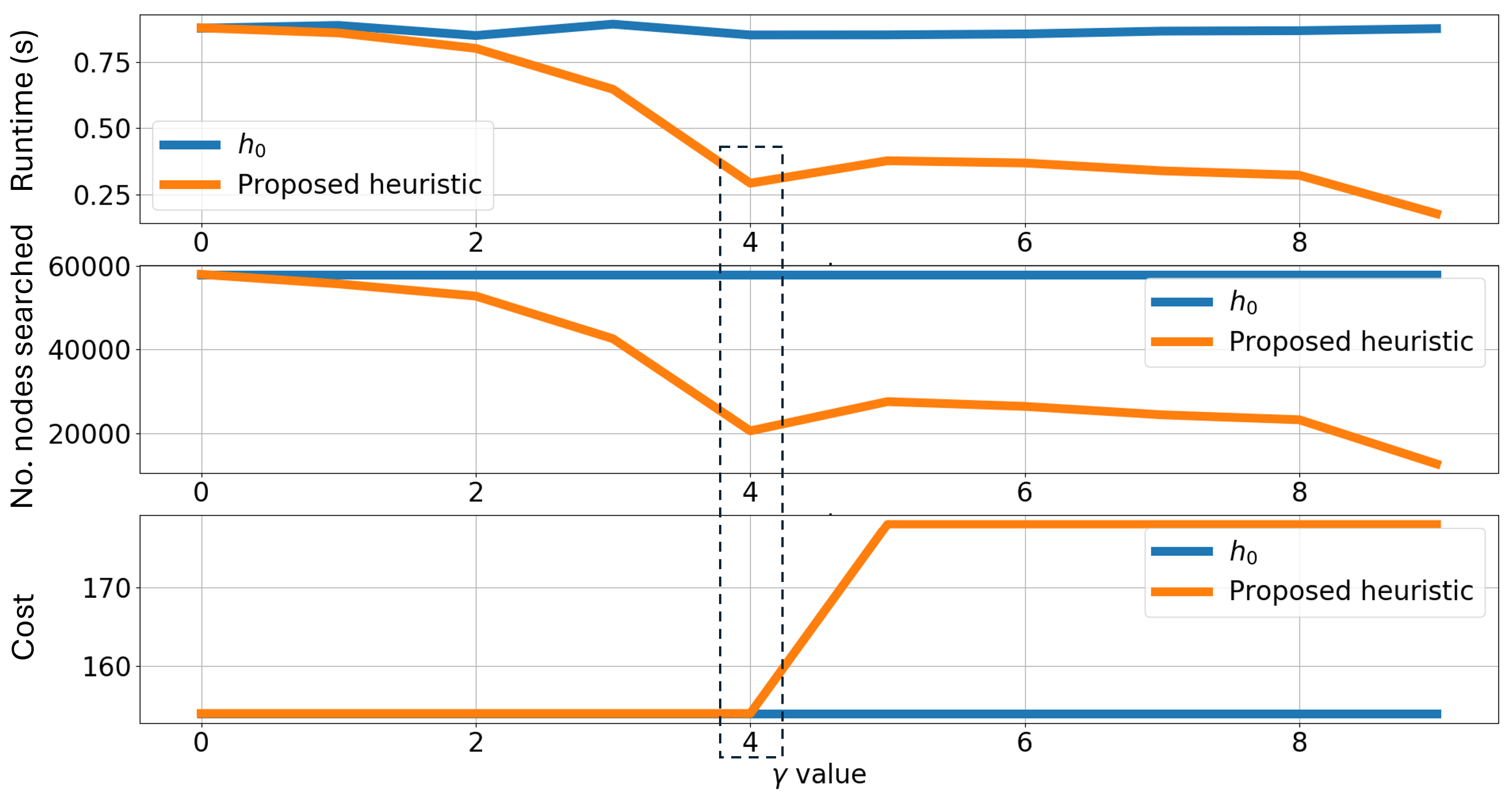}
    \caption{Empirical determination of the scaling factor}
    \label{fig:finding_gamma}
\end{figure}

\noindent\textbf{Empirically determining the scaling factor}. To determine the value of $\gamma$ for a given problem, we consider a fixed problem setup by keeping the environment size, labeled locations, specifications and relaxation preference unchanged across iterations. Thus, the same problem instance is solved for different values of $\gamma$. We assess the resulting cost, runtime as well as number of nodes searched in the solution space. Fig.~\ref{fig:finding_gamma} shows one such example for a grid environment of size (50,50) for a task of visiting 5 locations in no specific order, $\phi = \Event a \land \Event b \land \Event c \land \Event d \land \Event e$. We choose $\gamma = 4$ since it corresponds to an optimal cost while reducing the time and memory usage considerably. The choice of $\gamma$ is thus a design decision that balances the trade-off between computational cost and search efficiency.\\

\noindent\textbf{TL planning}. It is important to note that the potential sub-optimality of the proposed heuristic does not affect the precise satisfaction of the temporal logic goals. 
In this case study, we aim to compare the trajectories given by the proposed approach with a baseline case of zero heuristic (uninformed search) denoted by $h_0$. The baseline is guaranteed to find an optimal solution for the given environment and specification. However, the optimal solution may not be unique. 

Consider a 20x20 grid environment given as a weighted transition system $\TS$ with seven atomic propositions (labels) $AP = \{a,b,c,d,e,h,i\}$. Each transition has a unit weight and the self-loop transitions incur zero cost. The task specification is $\phi = (\Event \; a \land  (\Event  (b \land \Event \; c)) \land (\Event (d \land \Event \; e)) \land \Event \; h \land (\neg \; i \; \Until \;  h))$. In plain English, this translates to \textit{``At some point in the future, visit $a$ followed by $b$ then visit $c$. Visit $d$ followed by $e$. Visit $h$ and do not visit $i$ until $h$ is visited."} Notice that the specification does not impose any ordering on $a$, $d$ and $h$. Thus, there are multiple possible ways in which the specification can be satisfied in $\TS$.  

For evaluating our approach, we set $\gamma = 15$ and the initial state to $(0,0)$. The atomic propositions are then randomly assigned to the nodes in $\TS$. As shown in Fig.~\ref{fig:cs1}, our approach provides at least an order of magnitude speedup in runtime and reduction of nodes explored. As expected, in some cases, this speedup is achieved at a cost of optimality (Fig.~\ref{fig:cs1}(b)). Compared to $h_0$, our heuristic finds an optimal solution with 93.5\% reduction in time, 93.2\% reduction in memory for scenario 1 and a close-to-optimal solution for 94\% reduction in time, 93\% reduction in memory for scenario 2.

\noindent{\textbf{{Relaxation}}
For the same specification, if labels $b$ and $e$ are not present in the environment, the user preferences are (1) MRP: in place of $b$, $k$ can be visited with a penalty of $3$, (2) MRP: instead of $b$, $j$ can be visited while incurring a penalty of $5$, (3) MVP: visit to $e$ can be canceled with a penalty of 2. In other words, 1) $b \mapsto^3 k$, 2) $b \mapsto^5 j $, 3) $e \mapsto^2 \epsilon$. We set $\gamma=15$.  The resulting trajectories are shown in Fig.~\ref{fig:cs1_relaxation}. As depicted in scenario 1, the proposed heuristic finds an optimal path by substituting for $b$ with the lowest revision penalty and violating $e$ for the minimum violation penalty with 92.6\% and  92.13\% improvements in time and memory respectively. On the other hand, for scenario 2, our approach yields a sub-optimal path by using a relaxation preference with higher penalty resulting in a cost overhead of 16 while improving the time and memory usage by 91.8\% and 90.4\% compared to $h_0$. 


\begin{figure}
    \centering
    \includegraphics[width=.77\columnwidth]{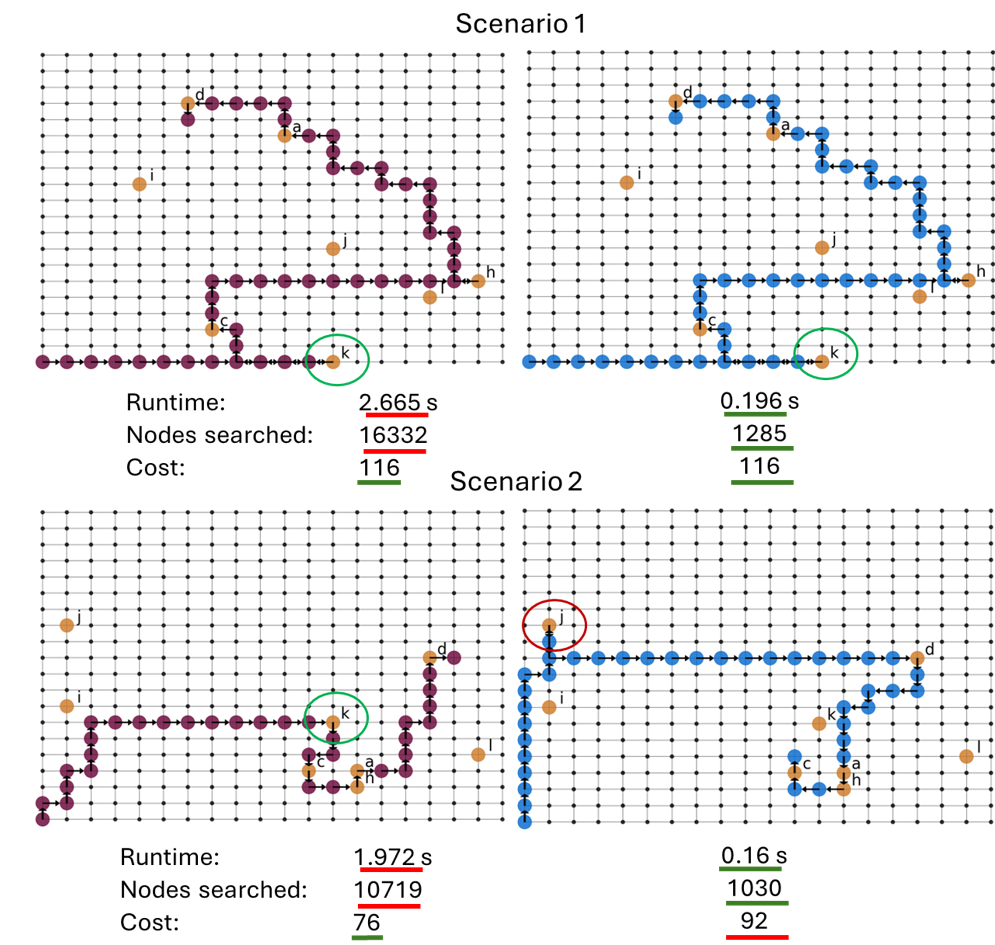}
    \caption{\scriptsize{Trajectories with relaxation obtained using baseline and the proposed heuristic. The green circles indicate satisfaction of a minimally relaxed subspecification. The red circle denotes a sub-optimal relaxation.}}
    \label{fig:cs1_relaxation}
\end{figure}

\subsection{Large Scale Robot Environments}
\begin{table*}[t]
\centering
{\scriptsize
\caption{\small{Comparison w.r.t. baseline for large scale robot environment}}
\label{table:scalability}
\begin{tabular}{|c|c|c|c|c|c|c|c|}
\hline
$\textbf{Specification}$ & \makecell{\textbf{Solution}\\{\textbf{Space Size}}} & $\BF{\vert \AP \vert}$ & $\gamma$ & \makecell{\textbf{Precomp.}\\ \textbf{Time (ms)}}& \textbf{Runtime (s)} & \makecell{\textbf{Nodes}\\{\textbf{searched}}} & \textbf{Cost}\\ \hline
$ \phi_1$  & nodes: 1512160 & 3 & - & - & 8.34 ($h_0$)  & 689193($h_0$) & 43151.37($h_0$) \\
 & edges: 11316640 &  & $\gamma_1$ = 27000 & 0.11 & \ul{3.83} & \ul{326860} & \ul{43151.37} \\ \hline 

& nodes: 4536480 & 4 & &  & 147($h_0$)  & 844181($h_0$) & 67546($h_0$)\\
$\phi_1^{relax}$  & edges: 45266560  & &  $\gamma_1 =27000 $ & 0.11 & \ul{114} & \ul{649860} & \ul{67546}\\ \hline 


    & nodes: 13609440 & 7  & -  & -  & 28($h_0$)  & 1643956($h_0$) & 28996($h_0$)\\
      $\phi_2$                             & edges: 298759296  &   &  $\gamma_{21} = 500 $ & 1.1 & \ul{10.8} & \ul{669695} & \ul{28996}\\
&                  &    &  $\gamma_{22}$ = 800 & 1.1 & \ul{6.58} & \ul{460877} & 33654\\  \hline 

                                & nodes: 40828320 & 9 & -  & -  & 233.8($h_0$)  & 1364344($h_0$) & 22746.3($h_0$)\\
              $\phi_2^{relax}$                           & edges: 1195037184  &   &  $\gamma_{21} = 500 $ & 1.9 & \ul{68.3} & \ul{611298} & 22766.5\\
               &                  &    &  $\gamma_{22}$ = 800 & 1.7 & \ul{44.98} & \ul{422999} & 29973.8\\  \hline

                     & nodes: 18145920  & 8  &  -                    & - & 237.17 ($h_0$)  & 13941839 ($h_0$) & 81844($h_0$) \\
 $\phi_3 $          & edges: 611098560 &    &  $\gamma_{31}$ = 2000  & 2.8 & \ul{53.04}     &\ul{3757966}    & 81850 \\ 
        &                  &    &  $\gamma_{32}$ = 5000 & 2.76 & \ul{16.87} & \ul{1216287} &  84446\\  \hline

                                & nodes: 54437760 & 9 & -  & -  & 2957.40($h_0$)  & 20333049($h_0$) &  78199($h_0$)\\
                                    & edges: 2444394240  &   &  $\gamma_{31} = 2000 $ & 4.73 & \ul{520.96} & \ul{4493701} & 94626.48\\
  $\phi_3^{relax}$                  &                  &    &  $\gamma_{32}$ = 5000 & 4.78 & \ul{269.6} & \ul{2507799} & 94626.48\\  \hline 
\end{tabular}}
\vspace{-3pt}
\end{table*}
To showcase the computational efficiency of the proposed heuristic, we use the New York City motorways network from OSMNX~\cite{Boeing2017} consisting of 378,040 nodes and 1,131,664 edges as $\TS$. We consider three representative cases: 1) sparse locations by considering only 3 labeled locations in the entire environment to be visited sequentially, 2) a complex grouping of tasks with choices, and 3) a complex sequential task. The scLTL specifications are:
{\small
\begin{enumerate}
    \item $\phi_1 = (\Event \;(groceries \;  \land \Event\;(fuel \; \land \; \Event \; bakery))) $
    \item  $\phi_2 = ((\Event \; lunch \; \land \; \Event (groceries  \; \lor \;  coffee)
     \land\;  \Event \; bakery) \lor (\Event \; fuel \; \land \; \Event (breakfast \; \land \;  (\Event \; bookstore))))$
    \item     $\phi_3 = (\Event \; lunch \; \land \; \Event (groceries \; \land \;\Event \; coffee)
     \land \;\Event \; bakery \land\; \Event \; (fuel \;\land \;\Event (breakfast \land \; \nonumber 
    (\Event \;bookstore)))\; \land \;(\neg\;rest \;U\; bakery)) $
\end{enumerate}}%
Additionally, the user preferences for relaxation in case of infeasibility are $fuel \mapsto^5 rest$ for $\phi_1$; $bakery \mapsto^5  mall, coffee \mapsto^3 lunch$ for $\phi_2$ and $\phi_3$. Table~\ref{table:scalability} summarizes the results. We denote by $\phi_i^{relax}$ the cases wherein relaxation is taken into account. The solution space size refers to the total number of nodes and edges given by $\vert X \vert \times \vert S_\FA \vert \times \vert Z_\ES \vert$ and $\vert \delta_\TS \vert \times \vert \delta_\FA \vert \times \vert \delta_\ES \vert$. It is crucial to note that we do not explicitly construct the solution space. $|\AP|$ denotes the number of locations allocated in the environment.  
As evident from these results, the proposed heuristic is simple enough to be pre-computed in a few milliseconds while drastically improving the search and runtime performance for a city-scale robot environment for complex tasks. Moreover, it can be seen that some values of $\gamma$ achieve the optimal cost in a fraction of time and memory as compared to the baseline ($\gamma_1, \gamma_{21}, \gamma_{31}$). Increasing this scaling factor further may improve the search time and memory considerably but may worsen the cost incurred ($\gamma_{22}$ and $\gamma_{32}$).

\begin{figure}
    \centering
    \includegraphics[width=\linewidth]{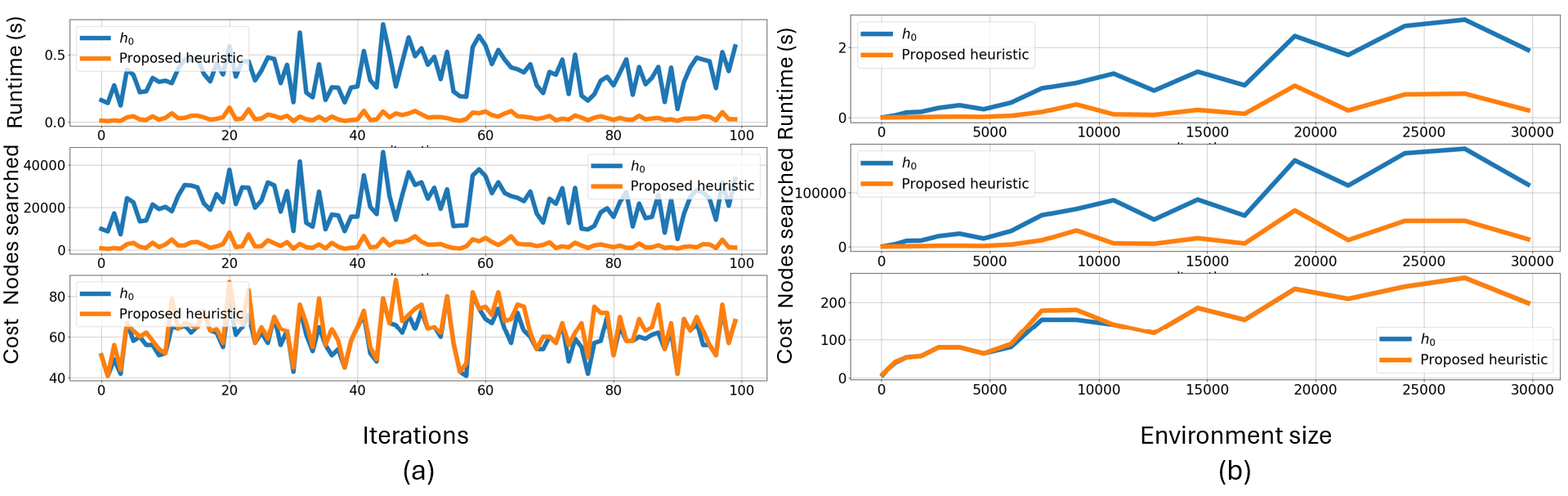}
    \caption{\scriptsize{ a) Effect of randomized AP assignment, b) Effect of varying $\TS$ size}}
    \vspace{-3pt}
    \label{fig:random_ap}
\end{figure}



\subsection{Runtime analysis}
\noindent\textbf{Randomized locations} For a 50x50 grid environment and $\phi = \Event a \land \Event b \land \Event c \land \Event d \land \Event e$, and $\gamma=4$, we vary the number of locations of each type (atomic proposition) and randomly assign minimum 1 and maximum 4 instances of each label. This gives rise to multiple possible paths that satisfy $\phi$ with varying path lengths. The results are shown in Fig.~\ref{fig:random_ap}(a). 
For some problem instances, our approach chooses the labeled states in the environment that can be reached faster, albeit at a slightly higher cost. 

\noindent\textbf{Environment size} For the same $\phi$ and $\gamma =10$, we vary the environment size from a 25 states to 30000 states. The scaling factor is substantially smaller due to simpler structure of the specification. The improvement in runtime (and reduction of nodes explored) increases consistently with the increasing environment size and is more pronounced for larger environments as shown in Fig.~\ref{fig:random_ap}(b). Notably, except for a few cases between $\vert X \vert = 5000 $ and  $\vert X \vert = 10000 $, the proposed heuristic achieves the optimal cost across all remaining instances.


\subsection{Empirical study on bounded suboptimality}
We consider a 100x100 grid environment wherein a robot is tasked to perform $\phi = (\Event a \land  (\Event (b \land \Event c)) \land (\Event (d \land \Event e)) \land \Event h \land (\neg i \Until h))$. Varying $\gamma$ from 0 to 30000, we compare the relative error $\Delta$ in cost $\hat{J}(\textbf{x})$ between the baseline and proposed approaches where $\Delta = (\hat{J}_{h} - \hat{J}_{h_0})/ \hat{J}_{h_0}$. The results are depicted in Fig.~\ref{fig:rel_error}. This study underscores that the suboptimality of the proposed heuristic is bounded and the approximate empirical bound is $  \hat{J}_{h} \leq 1.5 \hat{J}_{h_0}$.

\begin{figure}[t]
    \centering
    \vspace{-2pt}
    \includegraphics[width=0.68\linewidth]{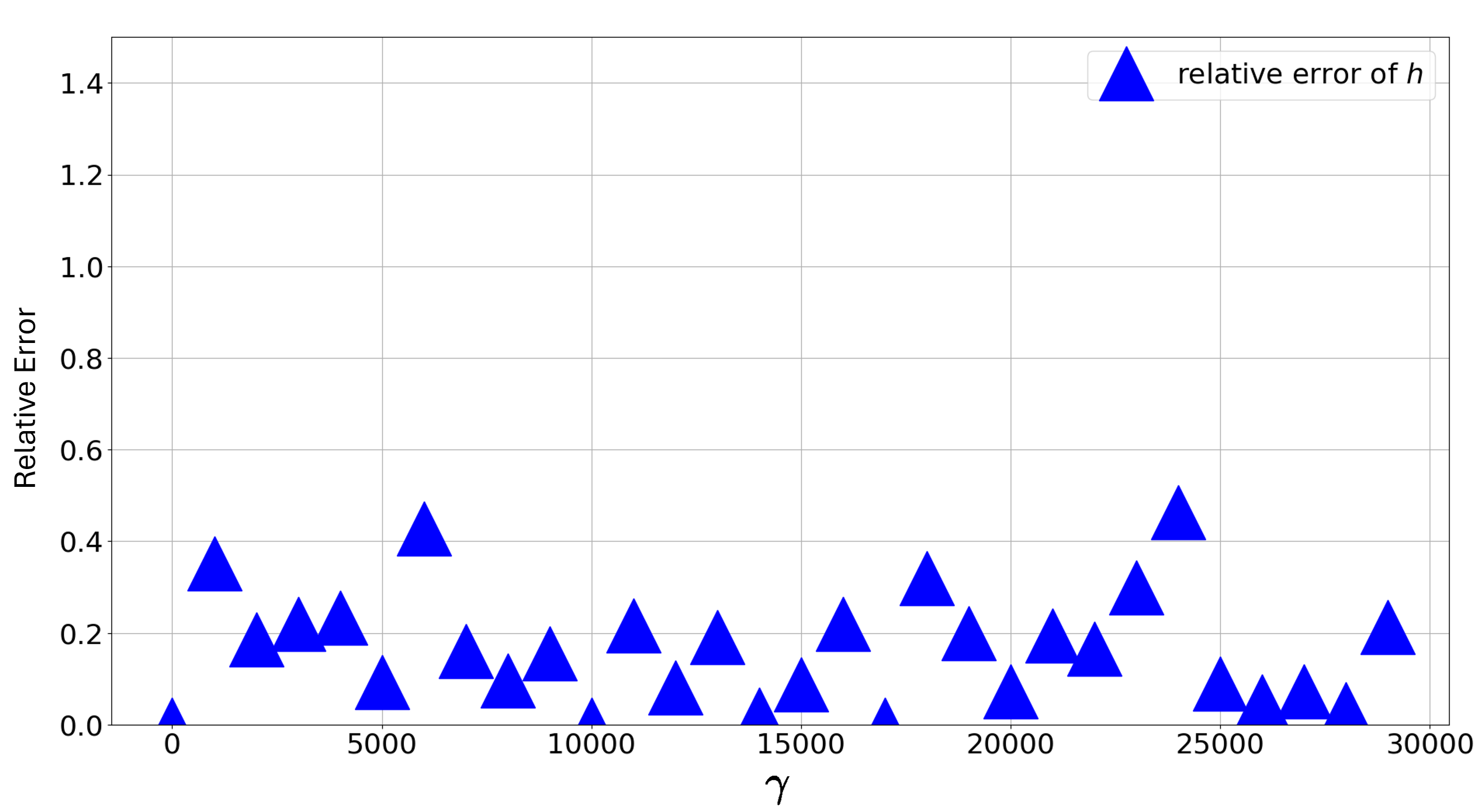}
    \caption{Relative error between the cost for the proposed heuristic and the optimal cost}
    \label{fig:rel_error}
\end{figure}

\section{Conclusion} 
This work presents an A$^{*}$-based search algorithm for planning for temporal logic tasks and user preferences for relaxation to address potential infeasibilities. The proposed approach avoids explicit product construction and instead implicitly searches through the reachable solution space. To facilitate this search, we propose an efficient, practicable heuristic that informs the search based on the distance from satisfaction with respect to the relaxed TL task. The proposed heuristic significantly reduces memory usage and runtime while achieving near-optimal costs. We provide runtime analysis and empirical suboptimality bounds. Future work will focus on a deeper investigation of the algorithm's theoretical properties and rigorous suboptimality bounds.

\bibliographystyle{IEEEtran}
\bibliography{relaxed_specs}
\end{document}